\pdfoutput=1

\documentclass[11pt]{article}

\usepackage{EMNLP2023}
\usepackage{graphicx}
\usepackage{xcolor}
\usepackage{amsmath}
\usepackage{times}
\usepackage{latexsym}
\usepackage{tikz}
\usepackage{bbm}
\usepackage[T1]{fontenc}

\usepackage[utf8]{inputenc}

\usepackage{microtype}

\usepackage{inconsolata}

\title{Evaluating Dynamic Topic Models}

\author{Charu James, Mayank Nagda, Nooshin Haji Ghassemi  \\ \textbf{Marius Kloft}, \textbf{Sophie Fellenz} \\
RPTU Kaiserslautern-Landau \\  Kaiserslautern, Germany\\  \texttt{surname}@cs.uni-kl.de }

\begin{document}
\maketitle
\begin{abstract}
There is a lack of quantitative measures to evaluate the progression of topics through time in dynamic topic models (DTMs). Filling this gap, we propose a novel evaluation measure for DTMs that analyzes the changes in the quality of each topic over time. Additionally, we propose an extension combining topic quality with the model's temporal consistency. We demonstrate the utility of the proposed measure by applying it to synthetic data and data from existing DTMs. We also conducted a human evaluation, which indicates that the proposed measure correlates well with human judgment. Our findings may help in identifying changing topics, evaluating different DTMs, and guiding future research in this area.
\end{abstract}

\section{Introduction}

Dynamic Topic Models (DTMs) \cite{blei2006dynamic} learn topics and their evolution over time from a time-indexed collection of documents. DTMs have been proven to be useful in various domains, including text mining \cite{mccallum2005author, wang2007topical, ramage2011partially, gerrish2011predicting}, computer vision \cite{fei2005bayesian, cao2007spatially, chong2009simultaneous}, and computational biology \cite{pritchard2000inference, zheng2006identifying}. DTMs enable summarization, browsing, and searching of large document collections by capturing the changes of topics over time. However, evaluating DTMs can be challenging due to their unsupervised nature, although it is crucial for effectively detecting trends in time-indexed documents.

Currently, the development of evaluation measures is not keeping pace with the advancements in new models\cite{doogan2021topic}. 
While traditional evaluation measures \cite{dieng2019dynamic,blei2006dynamic} can assess the quality and diversity of topics, they fail to capture the smoothness of topic changes over time. This limitation becomes problematic when a DTM exhibits high topic quality but lacks temporal smoothness. In such cases, existing evaluation measures may erroneously assign a high score to the model, even if there are rapid and abrupt transitions between topics. For instance, if a topic quickly shifts from ``politics'' to ``sports'', conventional evaluation measures might still rate the model positively. To accurately assess the quality of a DTM, it is crucial to consider the smoothness of topic changes over time, which can help identify gradual topic drifts or sudden shifts. Unfortunately, existing evaluation measures lack the capability to effectively track topic changes over time. To bridge this gap, we propose Temporal Topic Quality (TTQ)---a novel evaluation measure specifically designed for DTMs. TTQ incorporates the changes in topic quality into its assessment, thereby capturing the temporal characteristics of topics in DTMs.

\begin{figure*}[t]
\centering
\includegraphics[width=2\columnwidth]{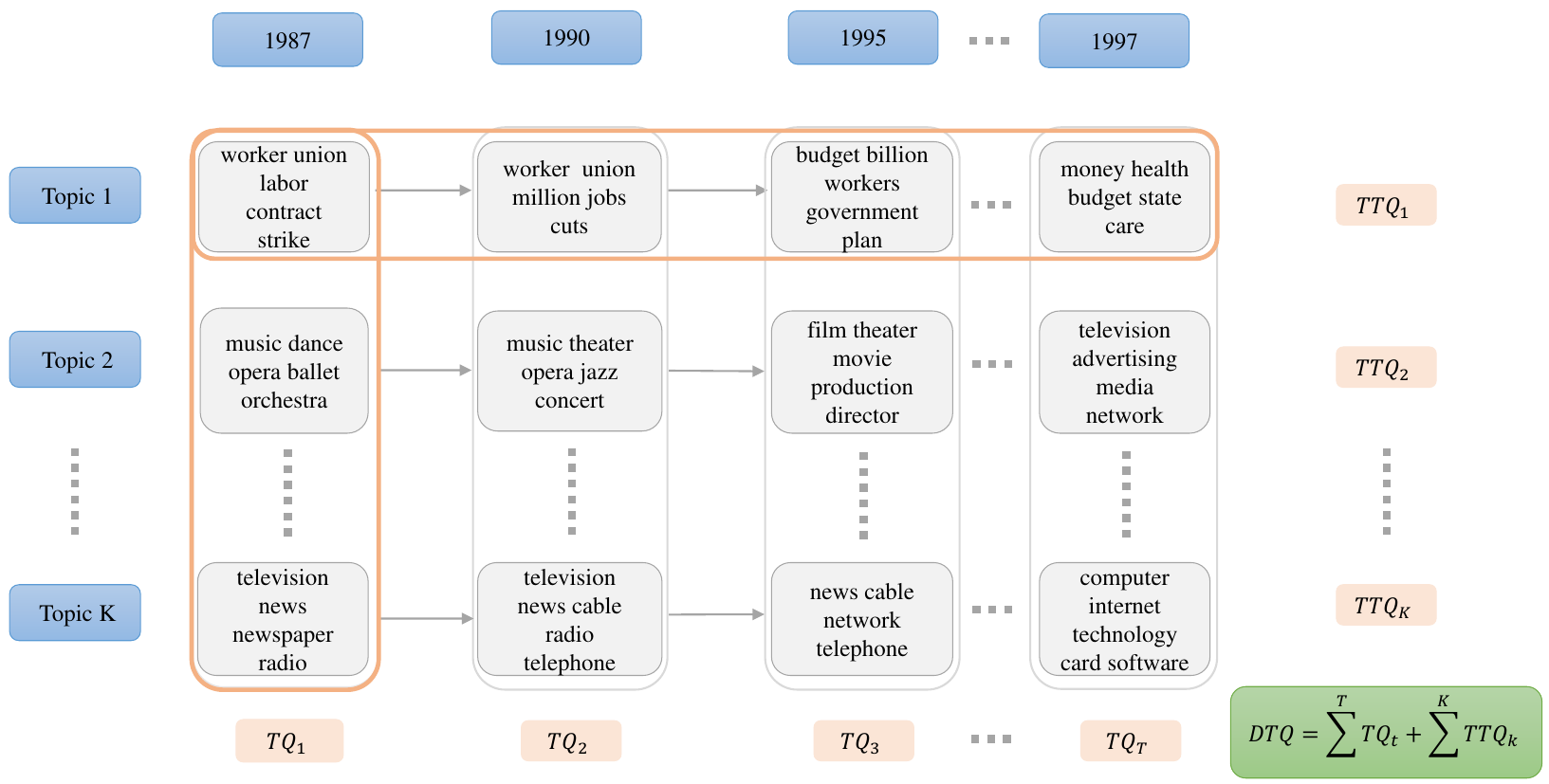} 
\caption{This figure illustrates the core concept presented in this paper. It illustrates the topic structure within DTMs. The vertical box highlights the set of topics for the first year, and the horizontal box shows the evolution of Topic~1 over time. Topic Quality (TQ) evaluates the topics for each year vertically, whereas Temporal Topic Quality (TTQ) evaluates each topic horizontally, capturing both the topic's evolution over time and the smoothness of topic progression.}
\label{dtm}
\end{figure*}

We provide empirical evidence of the effectiveness of the proposed measure by evaluating it on both synthetic and real topics. The results demonstrate a positive correlation between human ratings and the individual components of the TTQ measure. To provide an overall evaluation of DTMs, we propose the Dynamic Topic Quality (DTQ). The DTQ measure aggregates the TTQ measure with the static topic quality assessment. This aggregation is performed for both year-wise evaluations and temporal topic assessments, as illustrated in Figure~\ref{dtm}. In our experiments, we compare the results obtained using the DTQ measure with those obtained using previously employed measures for different topic models. The findings indicate that the DTQ measure provides a superior indication of the smoothness of topics in trained DTMs compared to the measures used in the past. We anticipate that the introduction of the new measure will contribute to improved comparisons between DTMs in future research endeavors. 
Our contributions can be summarized as follows:
\begin{itemize}
    \item We present a novel evaluation measure for DTMs that integrates both the vertical (year-wise) and the horizontal (temporal) dimension in the quality estimate (See Figure \ref{dtm}).
    \item We conduct a meta-analysis of prominent (statistical and neural) DTMs with our novel evaluation measures and present our findings.
    \item We show a positive correlation between human evaluations and the new evaluation measures, confirming their soundness.
\end{itemize}

\section{Related Work}

\label{section:related_work}
This section presents the previous work on DTMs and their evaluation approaches. We further discuss how the human evaluation for new measures was conducted in this domain.


\paragraph{Dynamic Topic Models} are developed to  model topics over time. Dynamic latent Dirichlet allocation (D-LDA)\cite{blei2006dynamic} extends the original LDA method\cite{blei2003latent} to account for temporal characteristics of sequential text data. The dynamic embedded topic model (D-ETM) combines D-LDA and word embeddings with a recurrent neural network (RNN)\cite{dieng2019dynamic}. A continuous-time version of DTMs was introduced by Wang \textit{et al.}~\cite{wang2008continuous,wang2006topics}. These models are not applicable to our datasets. Later work focused on the scalability of DTMs due to their computationally intensive training\cite{jahnichen2018scalable,bhadury2016scaling}. Many other DTMs have been proposed for different purposes, and not all are directly based on LDA \cite{grootendorst_bertopic_2022,he_dynamic_2013, zhou2017topic, gou2018constructing, satoshi, qiaozhu, ahmed, ahmed2, Dubey, wang2006topics}. It can be expected that with the recent advent of neural topic models, more DTM variants will be published in the future. In this work, we only compare D-LDA and D-ETM as the major proponents of the statistical and neural DTMs.
\paragraph{Evaluation Measures for Topic Models}

Previous work has focused on measures for static topic models such as the \emph{topic coherence} as measured by the normalized pointwise mutual information (NPMI) or $C_v$ score~\cite{newman2010automatic,mimno2011optimizing, roder2015exploring}. 

With the rise of neural topic models, local minima during training have become an issue, which may lead to component collapse ~\cite{burkhardt2019decoupling}. To capture such and other problems, different topic diversity, redundancy, or overlap measures have been introduced \cite{burkhardt2019decoupling, gui_neural_2019, dieng2020topic}.  Dieng \textit{et al.} combine topic diversity with coherence scores, resulting in \emph{topic quality} scores~\cite{dieng2020topic}, which is the basis of the definition of our temporal topic quality measure. 
\paragraph{Human Evaluation for Topic Models}
The concepts of coherence and interpretability are ``simultaneously important and slippery'' \cite{lipton2018mythos, hoyle2021automated}. A topic is coherent when a set of terms, viewed together, enables human recognition of an identifiable category \cite{hoyle2021automated}. \citet{doogan2021topic} define an interpretable topic as one that can be easily labeled and has a high level of agreement on its labels. In the case of DTMs, the topic coherence, interpretability, and smoothness across the temporal dimension are vital for its purpose and are the focus of our study.

There are two main ways to carry out human evaluation of topic models: topic  intrusion and topic rating. Both were developed specifically to account for the topic coherence in static topic models. In the topic rating task, humans are presented with a topic and are asked to rate it on a scale. Previously, authors have used ratings on a three-point ordinal scale \cite{hoyle2021automated,mimno2011optimizing,aletras2013evaluating}.
The rating task is not directly transferable to DTMs since we need to also rate how the topic changes over time. In the topic intrusion task, topics are chosen randomly, and one word in the topic is replaced with a word from another topic. Human evaluators are then asked to identify the intruder word\cite{hoyle2021automated}. Here, we extend and tailor both tasks to the temporal evaluation of DTMs.

\section{Background on Topic Evaluation Measures}
This section reviews the most common evaluation measures for topic models, which form the basis for our proposed measures: Topic coherence, diversity, and quality.

\subsection{Topic Coherence}

NPMI\cite{roder2015exploring} is the most commonly used coherence measure. For topic $k$, it can be computed as
\begin{equation}
\phi_{k}=  \sum^{N}_{j=2} \sum^{j-1}_{i=1} \frac{\log \frac{P\left(w^{(k)}_{i}, w^{(k)}_{j}\right)+\epsilon}{P\left(w^{(k)}_{i}\right) P\left(w^{(k)}_{j}\right)}}{-\log (P\left(w^{(k)}_{i}, w^{(k)}_{j}\right)+\epsilon)},
\label{eq:coherence}
\end{equation}
where $(w_1^{(k)},\dots,w_N^{(k)})$ is a list of the top $N$ words in topic $k$, and $P(w^{(k)}_i,w^{(k)}_j)$ is the probability of words $w^{(k)}_i$ and $w^{(k)}_j$ occurring together in a document, which is approximated by counting the number of documents where both words appear together, divided by the total number of documents~\cite{aletras2013evaluating}. A sliding window is used that determines the words to be considered at a time.  The $C_v$ score \cite{roder2015exploring} extends the NPMI by creating content vectors using co-occurrences of words, then calculating NPMI and cosine similarity between words.

\subsection{Topic Diversity}
\label{sec:diversity}

There exist three different approaches to measure diversity in topic models. The measure by Burkhardt and Kramer ~\cite{burkhardt2019decoupling}  takes into account \textit{how often} a word is repeated across topics and not only \textit{if} it is repeated. Additionally, it allows us to compute the diversity for individual topics and not just for the whole topic model. It computes the diversity for topic $k$ as $\mathrm{d}_{k,C} = 1 - \mathrm{r}_{k,C},$ 
where $r_{k,C}$ is the redundancy of topic $k$ with respect to the other topics $C=(v_1,\hdots,v_{k-1},v_{k+1},\hdots,v_K)$, where $v_i$ denotes the list of words for topic $i$ and can be obtained as follows
\begin{equation}
\mathrm{r}_{k,C} = \frac{1}{K-1} \sum^{N}_{i=1}\sum_{q\in C}\mathbbm{1}(w_{i}^{(k)},q),
\label{eq:diversity}
\end{equation}
where $\mathbbm{1}(w_{i}^{(k)},q)$ is one if the $i$th word of topic $k$, $w_{i}^{(k)}$, occurs in topic $q$ and otherwise zero, and $K$ is the number of topics.  Redundancy close to zero indicates that a topic has words that do not occur in any other topic,  and redundancy close to one indicates that most words in a topic also occur in (multiple) other topics. This is the primary measure used in the current work.

A related measure by Gui et al. \cite{gui_neural_2019} computes the Topic Overlapping (TO). A high value in $\mbox{TO}$ indicates that the associated words frequently appear across topics and can therefore be considered background words \cite{gui_neural_2019}. Dieng et al. \cite{dieng2019dynamic} proposed a third measure, which computes the topic diversity as the percentage of unique words in the top N topics. Having a diversity near zero indicates redundant topics. All three measures rank topics in the same order and thus lead to the same correlation values in our experiments.

\subsection{Topic Quality}
Topic quality is defined as a combination of coherence and diversity. A high diversity ensures that words across topics are different, and a high coherence ensures that words within topics are highly related, resulting in high-quality topics. While Dieng et al. \cite{dieng2019dynamic} used NPMI for coherence and their own diversity measure, the two components can be exchanged with different coherence and diversity measures. For $K$ different topics, it is computed as
 \begin{equation*}
\mathrm{TQ}=\frac{1}{K}\sum_{k=1}^{K} \phi_{k} \cdot d_{k,C},
\end{equation*}
where $\phi_k$ can be any coherence measure such as NPMI or $C_v$ score. 

We have now introduced the most common measures to evaluate topic models. None of them is suitable to evaluate temporal topic changes.  Our proposed measure introduced in the next section will address this gap.

\section{Proposed Measures}
 First, we present the temporal topic coherence and smoothness measures. Then we show how we can combine these two to obtain temporal topic quality, which measures the quality of topic transitions over time. This measure is then used in an aggregated measure, the dynamic topic quality, which evaluates both crucial aspects in DTMs: the quality of the topic model in each year and the quality of topic transitions over time. 


\subsection{Temporal Topic Coherence}

\begin{figure}
    \centering
    \includegraphics[width=0.8\columnwidth]{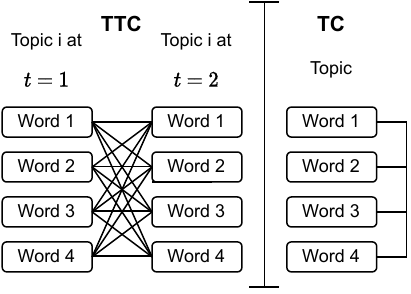}
    \caption{The idea of temporal topic coherence (TTC) in comparison with topic coherence (TC). TC considers word pairs within one topic. TTC only considers word pairs across time stamps of one topic.}
    \label{fig:ttc}
\end{figure}

Temporal topic coherence (TTC, see Fig. \ref{fig:ttc}) considers word pairs between two consecutive time stamps of one topic. Otherwise, the principle is the same as in TC: the co-occurrence of each word pair in the reference corpus is counted. Thus, if the topic stays the same semantically, TTC will be high, whereas if words of consecutive topics do not occur together in the reference corpus, TTC will be low.
More formally, we can now define the temporal topic coherence for window size $L$ as 
\begin{equation}
\begin{aligned}
& \text{TTC}_{k, t}= \\
& \sum_{j=1}^N \sum_{i=1}^N \frac{\log \frac{P\left(w_i^{(k, t)}, w_j^{(k, t+L-1)}\right)+\epsilon}{P\left(w_i^{(k, t)}\right) P\left(w_j^{(k, t+L-1)}\right)}}{-\log \left(P\left(w_i^{(k, t)}, w_j^{(k, t+L-1)}\right)+\epsilon\right)},
\end{aligned}
\end{equation}

where the variables are defined as in the definition of TC except $w_i^{(k,t)}$ is the $i$th word in topic $k$ and time stamp $t$.


\subsection{Temporal Topic Smoothness}
Topic diversity, introduced in Section \ref{sec:diversity}, measures how diverse topics are within one time stamp. The idea of temporal topic smoothness (TTS) is to use the diversity measure, but instead of applying it vertically for one topic model, we apply it horizontally over time (see Figure \ref{dtm}). In this case, the goal is to have smooth changes, which corresponds to a low diversity measure. Therefore, smoothness can be considered to be the opposite of diversity $d$. We apply TTS on one topic over a window of time steps.  TTS for topic $k$ in a window with size $L$ can be obtained by
\begin{align*}
  \text{TTS}_{k,t} =   r_{k,\tilde C},
\end{align*}
where $\tilde C=(v^{(k,t)},\hdots,v^{(k,t+L-1)})$ and $r_{k,\tilde C}$ is defined in Equation \ref{eq:diversity}. 

\subsection{Temporal Topic Quality}
Temporal Topic Coherence is calculated with respect to a reference corpus, whereas Temporal Topic Smoothness is only based on the words of the topics themselves. Thus, they are complementary since TTC can be high and TTS low or the other way around. High TTS and low TTC would point to component collapsing (an incoherent topic that is repeated over time), whereas high TTC and low TTS could point for example to changes in vocabulary use over time in topically coherent topics. As a combination of both measures, analogously to the topic quality measure, we propose the temporal topic quality (TTQ).

TTQ for topic $k$ over a sequence of time stamps $t=1,\cdots,T$ with a window size of $L$ can be computed as
\begin{align*}
\mathrm{TTQ}_k=
\frac{1}{T-L+1}\sum_{t=1}^{T-L+1} \text{TTC}_{k,t} \cdot \text{TTS}_{k,t}.
\end{align*}

The window size parameter $L$ enables to calculate the measure at different resolutions which correspond to detecting rapid changes (small window size) or slow transitions (large window size). 


\subsection{Dynamic Topic Quality}

TTQ enables us to see the changes of a topic over time (see Figure \ref{dtm} horizontal box). But it tells us nothing about the relations of different topics (for example, k=1 and 2) (see Figure \ref{dtm} vertical box). The role of TQ is to ensure the created topics are coherent and diverse. A DTM should exhibit high TTQ and TQ. Evaluating a model based on both measures gives rise to an aggregated measure called dynamic topic quality, $\mbox{DTQ}$, which measures the overall quality of a DTM and can be computed as
\begin{equation*}
  \mathrm{DTQ}=\frac{1}{2}\left[ \frac{1}{T}\sum_{t=1}^{T}\mbox{TQ}_t+\frac{1}{K}\sum_{k=1}^{K}\mbox{TTQ}_k \right],
\end{equation*}
where $\mbox{TQ}_t$ is the average quality of all topics in year $t$. Using $\mbox{DTQ}$, DTMs can be compared and ranked based on their performance on sequential text data.

\section{Using Human Evaluation to Examine DTMs}
\label{sec:Human_assessment}
\begin{figure}[t]
\centering
\includegraphics[scale=0.63]{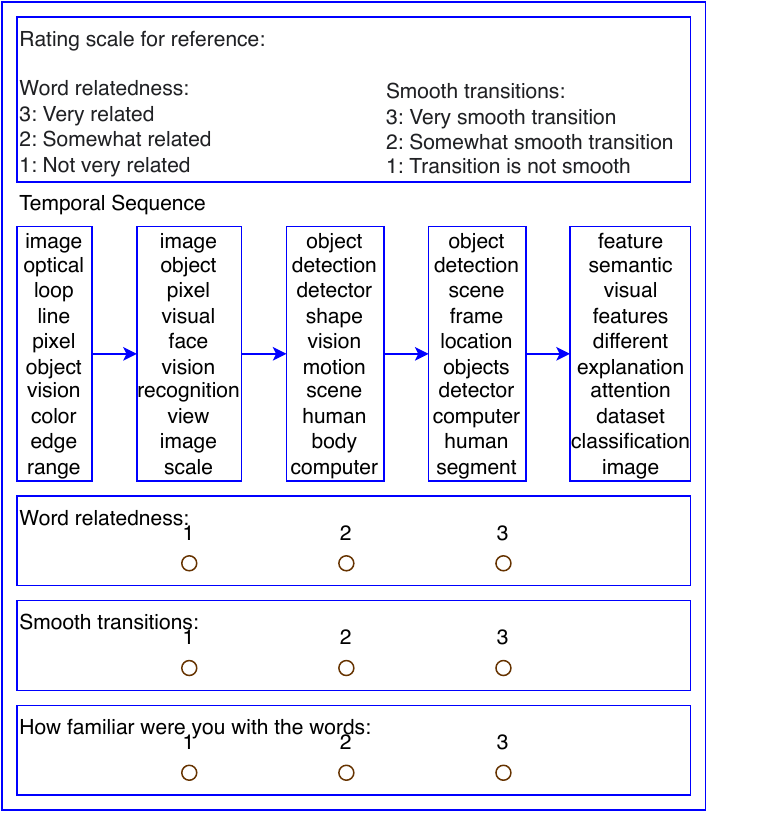} 
\caption{The temporal topic rating task as presented to the human evaluators. We present a sequence of five time stamps with equal spacing to evaluate a temporal topic.}
\label{fig:humanRatings}
\end{figure}

Since topic modeling is unsupervised and ground truth for topic quality is not available, we need to use human evaluation to validate our measures. In this section, we propose two tasks for the human evaluation of DTMs. The proposed tasks are adapted from the widespread \textit{word intrusion} \cite{chang2009reading} and \textit{topic rating} \cite{newman2010automatic} tasks of static topic models. 


\newcommand{\Arrow}[1]{%
\parbox{#1}{\tikz{\draw[->](0,0)--(#1,0);}}
}
\newcommand{\DownArrow}[1]{%
\parbox{#1}{\tikz{\draw[->](0,0)--(0,-#1);}}
}

\setlength{\tabcolsep}{3pt}
\begin{table*}[t!]
\begin{center} 
\begin{tabular}{l|ccc|ccc|c|ccc|ccc|c|}  
 & \multicolumn{7}{c}{\textbf{D-LDA}} & \multicolumn{7}{|c|}{\textbf{D-ETM}}  \\\cline{2-15}
 & \multicolumn{3}{c}{\textit{year-wise}} & \multicolumn{4}{|c|}{\textit{temporal} (ours)} & \multicolumn{3}{c}{\textit{year-wise}} & \multicolumn{4}{|c|}{\textit{temporal} (ours)} \\\hline
\textbf{Dataset} & TC & TD & TQ & TTC & TTS & TTQ & DTQ & TC & TD & TQ & TTC & TTS & TTQ & DTQ \\ 
\hline 
NeurIPS& 0.08  & 0.97 & 0.08  & 0.17 & 0.94  & 0.16 & 0.12  & 0.07 & 0.97 & 0.07 & 0.14 & 0.72 & 0.10 & 0.09\\
NeurIPS$^*$ & 0.08 & 0.97 & 0.08  & 0.00 & 0.07  & 0.00 & 0.04 & 0.07 & 0.97 & 0.07 & 0.00 & 0.06 & 0.00 & 0.03\\
\hline
NYT& 0.13 & 0.96 & 0.12  & 0.20 & 0.95  & 0.19 & 0.16 & 0.13  & 0.98 & 0.13 & 0.15 & 0.60 & 0.12 & 0.13\\
NYT$^*$ & 0.13 & 0.96 & 0.12 & 0.00 & 0.09 & 0.00 & 0.06 & 0.13 & 0.98 & 0.13 & 0.00 & 0.05 & 0.00 & 0.06 \\
\hline
UN Debates & 0.06 & 0.94 & 0.05  & 0.15 & 0.96  & 0.15 & 0.10  & 0.06 & 0.96 & 0.06 & 0.14 & 0.82 & 0.12 & 0.09\\
UN Debates$^*$ & 0.06 & 0.94 & 0.05 & 0.00 & 0.08 & 0.00 & 0.02 & 0.06 & 0.96 & 0.06 & 0.00 & 0.05 & 0.00 & 0.03\\
\hline
diff &\Arrow{.2cm}&\Arrow{.2cm}&\Arrow{.2cm}&\DownArrow{.2cm}&\DownArrow{.2cm}&\DownArrow{.2cm}&\DownArrow{.2cm}&\Arrow{.2cm}&\Arrow{.2cm}&\Arrow{.2cm}&\DownArrow{.2cm}&\DownArrow{.2cm}&\DownArrow{.2cm}&\DownArrow{.2cm}\\
\end{tabular} 
\end{center}
\caption{This table demonstrates that our temporal measures are able to capture temporal transitions, whereas the year-wise measures are not. NeurIPS$^*$ is a synthetic dataset where the original topics from the NeurIPS dataset are shuffled. On the shuffled topics, the temporal measures record lower scores as compared to the original topics, whereas the year-wise measures show unchanged values. This suggests that using only year-wise measures (TQ) to evaluate Dynamic Topic Models (DTM) is insufficient. The performance of D-LDA and D-ETM models are shown in terms of both \textit{year-wise} (TC, TD, TQ) and \textit{temporal} (TTC, TTS, TTQ)  measures. These measures are computed based on the NPMI scores on three real-world datasets of NeurIPS, NYT, and UN General Debates.}
\label{dtq}
\end{table*}

\subsection{Temporal Word Intrusion}
\label{sec:intrusion}
Word intrusion \cite{chang2009reading} is a common way of evaluating topic models. Intrusion words are chosen such that they have a low probability of belonging to the target topic but a high probability of belonging to a different topic. Words of existing topics are replaced by the intrusion words. Humans are then asked to detect the intrusion words. In temporal word intrusion, we instead modify a temporal sequence of one topic $k$, $(v_{k}^{(1)},\dots,v_{k}^{(T)})$, where $v_{k}^{(t)}$ corresponds to the list of top words for topic $k$ at time $t$. We can then analyze how our proposed measure for coherence over time changes based on the number of intruder words.\footnote{Note, that in contrast to \cite{hoyle2021automated} we do not use human evaluation here, but directly study correlation between intrusion level and coherence.}


\subsection{Temporal Topic Rating}
In the topic rating task, humans are presented with a topic and asked to rate it on a three-point ordinal scale.
Similarly, we aim to examine a topics' temporal sequence for \textit{word-relatedness} and \textit{smoothness} with the rating task. Human annotators are asked to rate a topic sequence instead of one static topic. Word familiarity scores are also collected for the analysis. Fig.~\ref{fig:humanRatings} shows how the task was presented to the human annotators.

\section{Experiments}

In this section, we establish the efficacy of the proposed measures using synthetic data and human evaluation. First, we compare static and temporal measures on synthetic topics in Section \ref{sec:temporal}. Second, we investigate the measures' sensitivity to noise in Section \ref{sec:intrusionAssessment}. Then, we conduct human evaluations of dynamic topics and compute the correlations of human ratings with the temporal measures in Section \ref{sec:human_eval}. As a window size parameter for our proposed measures TTS and TTC, we choose $L=2$ in all experiments. Choosing higher window sizes would make the measure more sensitive to detecting slower transitions. However, sudden changes are of greater interest to us as they affect the interpretability of a topic over time more. 

\subsection{Models and Corpora}
We compare our proposed measures for the models D-LDA\cite{blei2006dynamic} and D-ETM\cite{dieng2019dynamic}. D-LDA is a probabilistic model extending the popular LDA model to be dynamic. D-ETM is a neural DTM, which uses embeddings of words and topics. We use 50 topics as is common in the literature\cite{dieng2019dynamic,hoyle2021automated} for both the models. We randomly select 80\% of documents for training, 10\% for testing, and 10\% for validation.\\
We study our proposed evaluation measures using three commonly used datasets in the domain. The UN General Debates corpus \cite{DVN_0TJX8Y_2017} spans over 51 years (1970 to 2020). It contains general debate statements from 1970 to 2020. The second dataset \cite{AB2/GZC6PL_2008} consists of New York Times articles spanning over 21 years (1987 to 2007). The third dataset, the NeurIPS corpus \cite{swami_2020}, contains all NeurIPS papers from the period 1987 to 2019. Each dataset is preprocessed using standard techniques, such as tokenization and removing all punctuations and stop words (see Appendix \ref{sec:appendixA}, \ref{sec:appendixB}, and \ref{sec:appendixC} for complete corpus statistics,  preprocessing, and model training details). 

\subsection{Efficacy of Temporal Topic Evaluation} 
\label{sec:temporal}


Table \ref{dtq} compares the proposed temporal measures (TTC, TTS, TTQ, and DTQ) to static (year-wise) measures (TC, TD, and TQ).

\paragraph{Setup} 
We evaluate our proposed measures on three real-world corpora. Additionally, we construct synthetic topic models by shuffling the original topics of each model in each time stamp. This shuffling disrupts the topic transitions. The resultant synthetic data works as a proxy for the output of a DTM with poor topic transitions. An ideal evaluation measure is expected to capture the impact of shuffling on the topic transitions.



\paragraph{Results} Evaluation results for the shuffled topics (corpus with $^*$) are compared to results on unshuffled topics in Table \ref{dtq}. This reveals that the year-wise measures are unchanged irrespective of the shuffling. However, the temporal measures of the shuffled topics are significantly lower than the temporal measures of the original topics across all three datasets. This indicates that the proposed temporal measures can consistently detect poor topic transitions. The findings also stress that temporal changes are not reflected in the year-wise measures. Therefore, they are inadequate for evaluating dynamic topics.

Table~\ref{dtq} also shows that D-LDA and D-ETM have similar TQ, but exhibit different temporal behavior. D-LDA generally produces smoother topic transitions (higher TTS) than D-ETM. Furthermore, using the temporal evaluation measures, we can monitor changes in topics over time, as shown in Appendix \ref{sec:appendixD}. Table \ref{dtq-repeat} in Appendix \ref{sec:year-wise-efficacy} also shows why the combined measure DTQ is helpful to evaluate a DTM. It investigates the case where topic diversity is zero, which only affects DTQ, but not TTQ. 


\setlength{\tabcolsep}{3pt}
\begin{table}[t!]
\begin{center} 
\begin{tabular}{l|ccc|ccc}
 & \multicolumn{3}{c}{\textbf{D-LDA}} & \multicolumn{3}{|c}{\textbf{D-ETM}}  \\ \cline{2-7}
\textbf{Dataset} & \textbf{TTC} & \textbf{TTS} & \textbf{TTQ} & \textbf{TTC} & \textbf{TTS} & \textbf{TTQ} \\
\hline
NeurIPS & 0.97  & 0.98 & 0.98 & 0.95 & 0.91 & 0.91\\
NYT & 0.95  & 0.98 & 0.94 & 0.98 & 0.91 & 0.96\\
UN &0.87  & 0.99 & 0.94 & 0.89 & 0.92 & 0.88\\
\end{tabular} 
\end{center}
\caption{This table shows that all our measures correlate well with temporal word intrusion. Shown is the Spearman's correlation for temporal word intrusion for three datasets. All the correlations yield more than $95\%$ confidence intervals.}
\label{correlation-intrusion}
\end{table}

\subsection{Word Intrusion Assessment of Temporal Topics}
\label{sec:intrusionAssessment}

We investigate how our measure changes with respect to the intrusion level using the temporal word intrusion task presented in Section \ref{sec:intrusion}. We expect the temporal score to decrease with an increase in intrusion levels.

\paragraph{Setup} A target topic is randomly selected from each model and corpus. One time stamp is randomly chosen from the target topics' temporal sequence. Then, one randomly chosen word is replaced with a random intruder word in the selected topic. We repeat this process for ten intruder words. The number of intruder words determines the noise level chosen between one and ten.

An ideal temporal measure is expected to decrease with a higher intrusion level. To this end, we compute Spearman's correlation on a ranking of both intrusion levels and temporal measures for each model and dataset.

\begin{figure}[t] 
\centering
\includegraphics[scale=0.55]{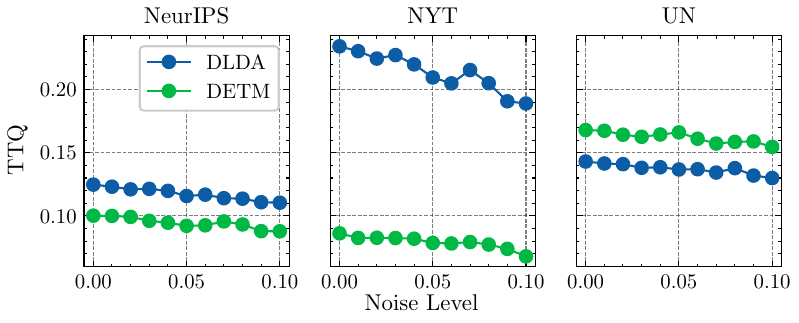}
\caption{Temporal intrusion detection. The figure shows temporal topic quality for different noise levels for one random topic (Topic 44) per dataset.  The TTQ measure decreases continuously as more intrusion words are added to the topics for all three datasets.}
\label{ttqvsnoise}
\end{figure}

\paragraph{Results} Table \ref{correlation-intrusion} presents the results for this experiment. Strong correlations for all datasets are achieved. This suggests that the proposed measures decrease with an increase in noise level.  Figure \ref{ttqvsnoise} shows this visually. Overall, the intrusion results underline the success of the proposed measures in differentiating between low- and high-quality topics and are sensitive to intrusion levels.

\subsection{Human Evaluation of Temporal Topics}
\label{sec:human_eval}

This section investigates the correlation between human evaluation scores and the proposed automated measures on a random sample of topics from each dataset. For this purpose, we use the temporal word-relatedness and smoothness tasks described in Section \ref{sec:Human_assessment} and shown in Figure \ref{fig:humanRatings}. 

\paragraph{Setup} We randomly select 20 topics from both models of each dataset for the human survey. For each topic, we present human raters with a sequence of five evenly spaced time steps. We conduct a separate study for each data corpus. We recruit crowd workers from Prolific.co and compensate them with 12 USD/hr equivalent. Every participant is provided with detailed instructions. We follow the protocol by \citet{hoyle2021automated} and recruit a large number of crowd workers (18) per task to ensure adequate statistical power. Aggregated human ratings of word relatedness and smoothness for each topic are calculated by averaging across all valid respondents.

We follow two criteria to identify valid respondents. The first one is based on a control task. Respondents who fail the control task are excluded from the analysis. Topics in the control task are created synthetically by randomly selecting words, resulting in a very low-quality topic in terms of word relatedness and smooth transitions. Second, we monitor the duration taken to complete the survey. We filter out outlier respondents based on the median time taken to complete the survey. These criteria for filtering out invalid respondents are consistent with the previous studies in the domain \cite{hoyle2021automated, chang2009reading}.


\paragraph{Baseline Comparisons} No standard baseline exists for temporal topic evaluations. Our extensive survey only found a few instances with a reference and motivation of temporal assessment of topics \cite{blei2006dynamic,dieng2019dynamic}. These references are limited to qualitative evaluations and deliver no quantitative measures.

Following the existing evaluation measures and taking inspiration from the references concerning the temporal assessment of dynamic topics in the literature, we derive baselines for topic coherence (B-TC) and topic smoothness (B-TS). The baseline measure for coherence is computed from the average score for topic coherence over time. For smoothness, we calculated $1 - \text{diversity}$ of one topic over time as the baseline which corresponds to TTS with maximum window size.


\begin{figure}[t!]
\centering
\includegraphics[scale=1.0]{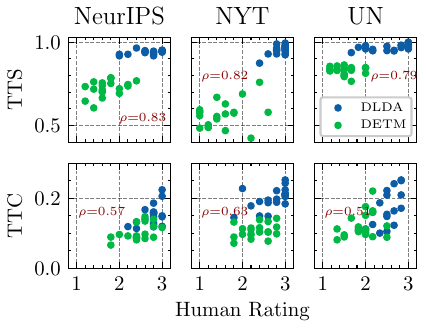} 
\caption{This figure shows that the temporal topic smoothness and coherence correlate well with the human evaluation. Correlation (Spearman’s $\rho$) between human and temporal topic smoothness (top) and coherence (bottom) for random topics from NeurIPS (left), NYT (middle) and UN (right) datasets.}
\label{scatter-human-vs-auto}
\end{figure}



\paragraph{Results} 
The scatter plots in Figure \ref{scatter-human-vs-auto} further show the correlation of human ratings and temporal measures for word-relatedness (bottom) and smoothness (top) respectively. Inspection of outlier points reveals that low human ratings and high temporal scores often belong to topics with low familiarity among raters. The figure shows that humans rated D-LDA topics higher as compared to D-ETM. It also shows that human ratings are more varied as compared to the automated measures.

Table \ref{correlation-ratings} shows each dataset's estimated Spearman's correlation. The coherence measures are correlated with the word-relatedness ratings and the smoothness measures are correlated with human smoothness ratings respectively. The proposed temporal measures consistently show a  strong correlation with human ratings  among all the datasets. All correlations obtained are in $95\%$ confidence intervals. The correlation scores align with the previous studies for B-TC \cite{hoyle2021automated}. 

Table \ref{correlation-ratings} shows that TTC correlates better with the human perception of word relatedness in DTMs than the baseline TC which does not consider temporal transitions. The  TTS also correlates well. The baseline topic smoothness also has a high correlation with human-perceived smoothness which is to be expected since it corresponds to TTS with maximum window size. This suggests that, depending on the dataset, the TTS measure is fairly robust with respect to the window size.

Figures \ref{box-plot-c} and \ref{box-plot-s} in Appendix \ref{sec:appendixE} show the results of the human and automated evaluation for word-relatedness and smoothness, respectively. These figures indicate that human ratings align with our proposed temporal measures.

\setlength{\tabcolsep}{6pt}
\begin{table}[t!]
\begin{center} 
\begin{tabular}{l|cc|cc}
\textbf{Dataset} & \multicolumn{1}{c}{\textbf{TTC}} & \multicolumn{1}{c}{\textbf{B-TC}}  & \multicolumn{1}{|c}{\textbf{TTS}}& \multicolumn{1}{c}{\textbf{B-TS}}\\ 
\hline
NeurIPS &\textbf{0.57} & 0.21 &\textbf{0.83} &0.65\\
NYT &\textbf{0.63} & 0.17 & 0.82 &\textbf{0.83}\\
UN &\textbf{0.51} & -0.02 & 0.79 &\textbf{0.81}\\
\end{tabular} 
\end{center}
\caption{This table shows that 1) TTC correlates better with human evaluation than B-TC 2) TTS correlates better or to a similar degree as B-TS with human smoothness evaluation. Spearman's correlation coefficients between mean human evaluation and automated measures. These correlations of proposed measures (TTC and TTS) are compared to the baseline measures (B-TC and B-TS). The highest correlations for each pair are shown in bold.} 
\label{correlation-ratings}
\end{table}

\section{Conclusion}
\label{section:conclusion} 

This paper fills a gap in evaluating temporal characteristics of DTMs. We complement the existing year-wise measures by proposing novel temporal measures. Our proposed temporal measures capture different aspects of temporal topic changes in DTMs. We show that our measures are able to better capture temporal characteristics of topic changes than their year-wise counterparts and have positive correlation with human evaluations. We show the efficacy of our measure by evaluating two dynamic topic models and demonstrating their different temporal characteristics. Our proposed evaluation measures improve future comparisons between DTMs.

\section*{Limitations}

We identify two main limitations of our approach. The first concerns the human evaluation. Here, we have to rely on the quality of answers by human annotators. Although we took care of recruiting a large number of annotators (18) in order to lower the variance of our results, it would be preferable to have fewer annotators who provide high-quality annotations. This could only be achieved by training people before they are given the task which would require a training protocol. This is outside the scope of our study. This issue of human annotators not being experts in the dataset domain or in the given task affects other studies as well and is difficult to solve. Automated measures have to be validated using human annotations, but human annotations are never perfect. Therefore, there will always be a gap remaining.

The second limitation of automated topic evaluation, in general, is the reference corpus. Automated measures are calculated with respect to a reference corpus. If words or topics are not present in the reference corpus (for whatever reason), the result will be suboptimal. This is especially true for temporal topics where the number of documents for selected time steps can be small which might lead to the respective topics not being present in the reference corpus. This could be addressed in the future by selecting the reference corpus more carefully or potentially choosing an external reference corpus.

\section{Acknowledgements}
The authors was funded by the German Federal Ministry of Education and Research under grant number 01IS20048. Responsibility for the content of this publication lies with the author. Additionally, we acknowledge support from the Carl-Zeiss Foundation and the DFG awards BU 4042/2-1 and BU 4042/1-1.

\bibliography{anthology,custom}
\bibliographystyle{acl_natbib}

\appendix

\section{Dataset Details}
\label{sec:appendixA}
We study our evaluation measure using three different datasets: NeurIPS, New York Times, and UN General Debates.  The table~\ref{Corpus_Statistics} shows corpus statistics for the datasets.
\setlength{\tabcolsep}{2pt}
\begin{table}[hbt]
\begin{center} 
\begin{tabular}{l|rrr}
\textbf{Dataset} & \multicolumn{1}{c}{\textbf{NeurIPS}} & \multicolumn{1}{c}{\textbf{    NYT}}  & \multicolumn{1}{c}{\textbf{UN}}\\ 
\hline
Domain& \multicolumn{1}{c}{Science}  & \multicolumn{1}{c}{News}  & \multicolumn{1}{c}{Politics}\\
Number of Docs & {9,679} & {274,665}  & {8,481}\\
Vocab Size & 3,102  & 8,240 & 3,005\\
\end{tabular} 
\end{center}
\caption{Corpus Statistics. Shows the dataset that varies in the domain, document number, and vocab size. The NeurIPS is from \cite{swami_2020}, NYT is from \cite{AB2/GZC6PL_2008} and UN General Debates is from ~\cite{DVN_0TJX8Y_2017}.}
\label{Corpus_Statistics}
\end{table}

\section{Preprocessing Details}
\label{sec:appendixB}
The datasets undergo a series of preprocessing steps, which include converting the text to lowercase, eliminating stopwords, and removing punctuation marks. Tokenization is performed using Spacy \cite{spacy2}, and further refinement involves removing words that appear in less than a specified percentage (min\_df) of the documents, as well as words that occur in more than a specified percentage (max\_df) of the documents, using count vectorizer. The min\_df for the New York Times (NYT) dataset was determined to be 0.3\% after observing that a min\_df of 5\% yielded a vocabulary size of 564, which was considered insufficient for a large dataset. Table ~\ref{countVectorizor} shows the cut-off parameters used for the different datasets.  

\begin{table}[hbt]
\begin{center} 
\begin{tabular}{l|rrr}
\textbf{Dataset} & \multicolumn{1}{c}{\textbf{min\_df}} & \multicolumn{1}{c}{\textbf{    max\_df}}  & \multicolumn{1}{c}{\textbf{vocab size}}\\ 
\hline
NeurIPS& {5}\%  & {95}\%  & {3,102}\\
UN Debates & {5}\% & {95}\%  & {3,005}\\
NYT & 0.3\%  & 95\% & 9,046\\
\end{tabular} 
\end{center}
\caption{The table shows the vocab size that results from removing words that occur in less than min\_df percent of the documents and words that occur in more than max\_df percent of the documents.}
\label{countVectorizor}
\end{table}

\section{Model Training Details}
\label{sec:appendixC}
Expanding subsection 6.1, here we explain how hyperparameters are set for each topic model.

\textbf{D-LDA} For training the model, we use the Gensim python wrapper for dynamic topic models (DTM). We slice all datasets by year. As a result, every time slice contains all documents from that year. For each dataset, we ran 50 iterations with an alpha value of 0.01, which is a hyperparameter affecting the sparsity of the document topics for each time slice in the LDA models. In addition, we used top\_chain\_var values of 0.005.

\textbf{D-ETM} Using the skip-gram model, a 300-dimensional word embedding is obtained ~\cite{mikolov2013distributed}. The batch size for all datasets was 100 documents. We used the perplexity score on the validation set as stopping criteria for all datasets. The learning rate is set to a default value of 0.001. The hyperparameters delta, sigma, and gamma in D-ETM are set to 0.005 as suggested by the authors. A random selection of 80\% of documents is used for training, 10\% for testing, and 10\% for validation.

\section{Qualitative results}
\label{sec:appendixD}
\begin{figure*}[htbp]
\includegraphics[scale=0.12]{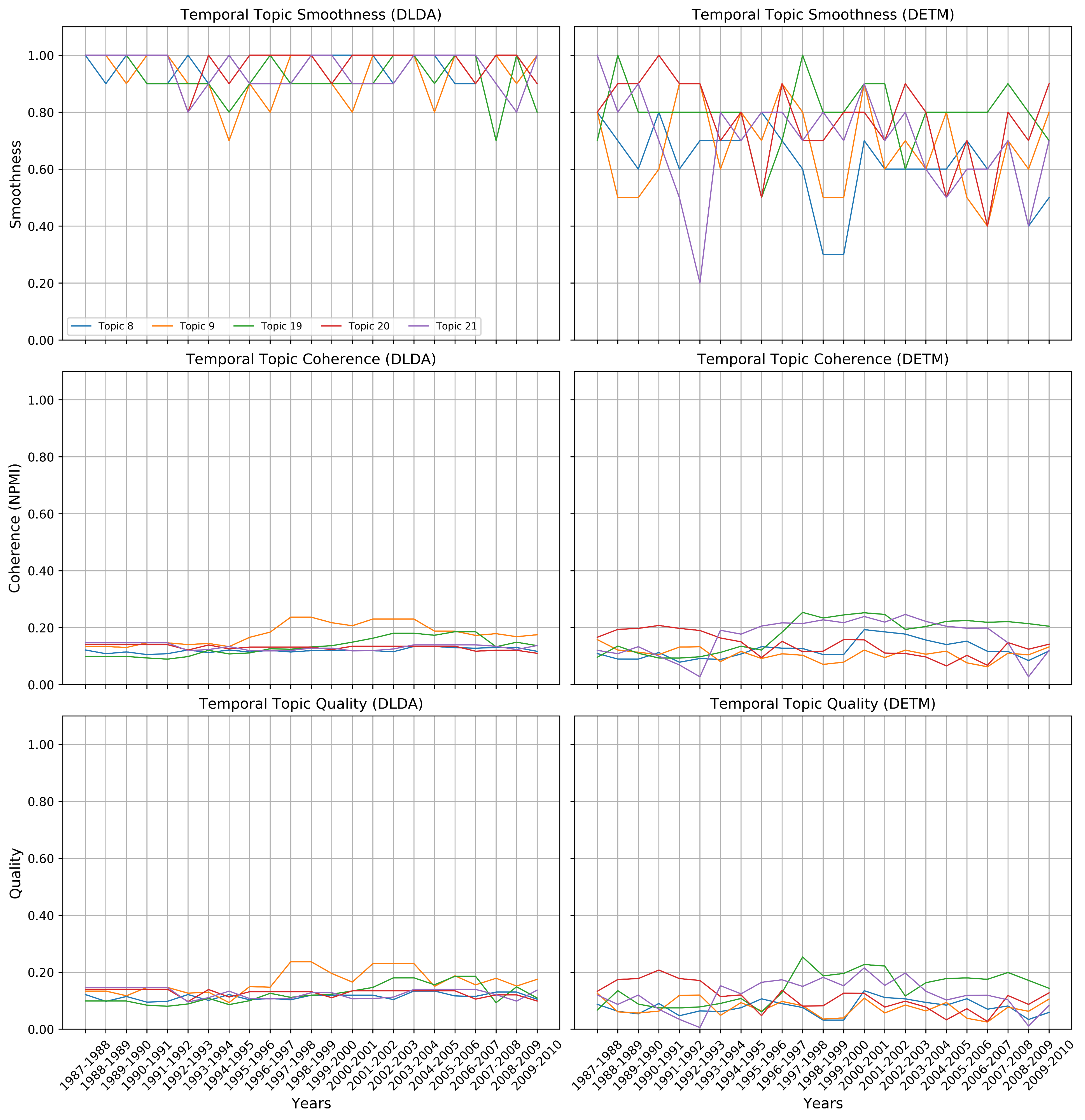} 
\caption{Shows how \textbf{temporal topic smoothness} changes over year for five topics generated by D-LDA (top-left) and D-ETM (top-right) for NeurIPS dataset.
Shows how \textbf{temporal topic coherence} changes over year for five topics generated by D-LDA (middle-left) and D-ETM (middle-right) based on \textbf{NPMI score}. Shows how \textbf{temporal topic quality} (TTQ) changes over year for five topics generated by D-LDA (bottom-left) and D-ETM (bottom-right).}\label{fig:DLDA_DETM_TTC_TTS}
\end{figure*}

A topic's temporal topic quality is determined by how smoothly its words change over time. The temporal topic quality is calculated in terms of temporal topic coherence (TTC) and temporal topic smoothness (TTS).
The concept of TTS is shown in the top row of Figure~\ref{fig:DLDA_DETM_TTC_TTS}. Topic 19 (top-left) illustrates an instance of topic words exhibiting smooth transitions. During the year 1988-1989, the TTS remains 1.0, indicating a lack of change in the topic words. Table~\ref{tab6}  provides a depiction of Topic 19 during this time frame. Furthermore, in the scenario where consecutive TTS score reach 1.0, the TTC score remains unchanged, as the topic words have not undergone any changes. 

Table~\ref{tab7} provides empirical evidence within the D-ETM model, demonstrating that Topic 8 experienced a relatively low TTS score between 1998-1999. Notably, despite the decline in TTS score, the corresponding topic remained largely unchanged, as indicated by the nearly same TTC score that did not exhibit a significant decrease. The same analysis applies to Topic 21, which represents a topic with drastic changes in its words. The TTS and TTC scores are observed to be low between 1992-1993, indicating a radical shift in the topic. Table \ref{tab8} shows topic 21 during this time. As the table shows there is a change from the topic of rule extraction to a topic on image object recognition at this point in time. This change can also be seen in temporal topic quality (TTQ) in Figure~\ref{fig:DLDA_DETM_TTC_TTS} wherein 1992-1993 the TTQ score is low.

The temporal topic coherence is calculated based on NPMI for D-LDA and D-ETM. Whereas D-LDA in general shows relatively unchanged temporal topic coherence over time, D-ETM exhibits more variance in TTC.

\setlength{\tabcolsep}{2pt}
\begin{table}[t]
\begin{tabular}{llll}
Year & words in Topic~19   & TTS & TTC \\
\hline
1988 & connectionist human figure    & 1.0 & 0.098\\
& systems research science   & \\
& knowledge performance  \\
& target rules \\
1989 & connectionist figure human &  1.0 & 0.098\\
&  rules knowledge target  & \\
&  performance science research& \\
& systems \\
1990 & figure connectionist rules  & 0.9 & 0.093\\
& human target knowledge   & \\
& performance science research & \\
& information \\

\end{tabular}
\caption{Topic 19 from D-LDA model using NeurIPS dataset, which shows the smoothness in topic during the year 1988-89, when TTS in Figure \ref{fig:DLDA_DETM_TTC_TTS} is 1.0, which is between the current year and previous year.}
\label{tab6}
\begin{tabular}{llll}
Year & words in Topic~8 & TTS & TTC  \\
\hline
1996 & position hand task user & 0.7 & 0.127 \\
& location based object &\\
&body target robot& \\
1997 & object position hand task & 0.6 & 0.126  \\
& robot user direction   &\\
&location right coordinates& \\
1998 & position hand human line  & 0.3 & 0.105 \\
& movement direction motor  & \\
& task object location & \\
1999 & spatial localization location & 0.3 & 0.104   \\
& light position human temporal  & \\
& subjects robot subject 

\end{tabular}
\caption{Topic~8 from the D-ETM model using the NeurIPS dataset, which shows low smoothness in 1998--1999, but TTC remains nearly the same. This is shown in Figure \ref{fig:DLDA_DETM_TTC_TTS} where the TTS score is 0.3 which is between the current year and previous year.}
\label{tab7}
\end{table}

\setlength{\tabcolsep}{3pt}
\begin{table*}[ht]
\begin{center} 
\begin{tabular}{l|ccc|ccc|c|ccc|ccc|c|}  
 & \multicolumn{7}{c}{\textbf{D-LDA}} & \multicolumn{7}{|c|}{\textbf{D-ETM}}  \\\cline{2-15}
 & \multicolumn{3}{c}{\textit{year-wise}} & \multicolumn{4}{|c|}{\textit{temporal} (ours)} & \multicolumn{3}{c}{\textit{year-wise}} & \multicolumn{4}{|c|}{\textit{temporal} (ours)} \\\hline
\textbf{Dataset} & TC & TD & TQ & TTC & TTS & TTQ & DTQ & TC & TD & TQ & TTC & TTS & TTQ & DTQ \\ 
\hline 
NeurIPS& 0.08  & 0.97 & 0.08  & 0.17 & 0.94  & 0.16 & 0.12  & 0.07 & 0.97 & 0.07 & 0.14 & 0.72 & 0.10 & 0.09\\
NeurIPS$^*$ & 0.05  & 0.00 & 0.00  & 0.13 & 0.94  & 0.12 & 0.06  & 0.17 & 0.00 & 0.00 & 0.11 & 0.78 & 0.08 & 0.04\\
\hline
NYT& 0.13 & 0.96 & 0.12  & 0.20 & 0.95 & 0.19 & 0.16 & 0.13  & 0.98 & 0.13 & 0.15 & 0.60 & 0.12 & 0.13\\
NYT$^*$ & 0.18 & 0.00 & 0.00  & 0.21 & 0.94  & 0.20 & 0.10 & 0.19  & 0.00 & 0.00 & 0.09 & 0.71 & 0.07 & 0.03\\
\hline
UN Debates & 0.06 & 0.94 & 0.05  & 0.15 & 0.96  & 0.15 & 0.10  & 0.06 & 0.96 & 0.06 & 0.14 & 0.82 & 0.12 & 0.09\\
UN Debates$^*$ & 0.01 & 0.00 & 0.00  & 0.06 & 0.95  & 0.06 & 0.03  & 0.12 & 0.00 & 0.00 & 0.09 & 0.79 & 0.08 & 0.04\\
\hline
diff &\Arrow{.2cm}&\DownArrow{.2cm}&\DownArrow{.2cm}&\Arrow{.2cm}&\Arrow{.2cm}&\Arrow{.2cm}&\DownArrow{.2cm}&\Arrow{.2cm}&\DownArrow{.2cm}&\DownArrow{.2cm}&\Arrow{.2cm}&\Arrow{.2cm}&\Arrow{.2cm}&\DownArrow{.2cm}\\
\end{tabular} 
\end{center}
\caption{This table demonstrates the need of having year-wise measures in the DTQ. We construct synthetic datasets (marked with $^*$) where we randomly select a topic and repeat it. The synthetic topics now work as proxy for the extreme case of component collapse in case of DTMs. On the synthetic topics, the year-wise measures record lower TQ scores as compared to the original topics, whereas the temporal measures show similar values.}
\label{dtq-repeat}
\end{table*}

\begin{table}[t!]
\begin{tabular}{llll}
Year & words in Topic~21   & TTS & TTC \\
\hline
1990 & rules rule cell extraction   & 0.7 & 0.099\\
& group clustering groups  & \\
& cluster expert clusters\\
1991 & rules rule extraction  &  0.5 & 0.068\\
& extracted group expert & \\
& groups clustering & \\
& induction self &\\
1992 & rules rule children expert  & 0.2 & 0.027\\
& extraction features view    & \\
& self image feature & \\
1993 & image surface view & 0.8 & 0.190\\
& recognition object matching & \\
& images correspondence views  &\\
& objects&

\end{tabular}
\caption{Topic 21 from D-ETM model using NeurIPS dataset, which shows the change in topic during the year 1992-1993, when TTS in Figure \ref{fig:DLDA_DETM_TTC_TTS} is 0.20 and TTC is 0.027 which is between the current year and previous year.}
\label{tab8}
\end{table}

\section{Efficacy of year-wise Topic Evaluation}
\label{sec:year-wise-efficacy}
In this section, we establish the efficacy of year-wise topic evaluation measures for evaluating DTMs. To this end, we construct synthetic topics which behave as proxies to output of a poor topic model. For the synthetic topics, we randomly select one topic from each model and dataset and repeat it while removing all other topics. The synthetic topics work as an extreme case of component collapse in a DTM. The results of this experiment are shown in Table \ref{dtq-repeat}. For all the synthetic versions of the three datasets, year-wise TQ is zero (because TD is zero). Hence, the overall DTQ is low as compared to TTQ. This establishes the efficacy of combining both TTQ and TQ in the form of DTQ when evaluating DTMs.

\section{Human Evaluations}
\label{sec:appendixE}

\begin{figure}[hbt]
\centering
\includegraphics[scale=0.55]{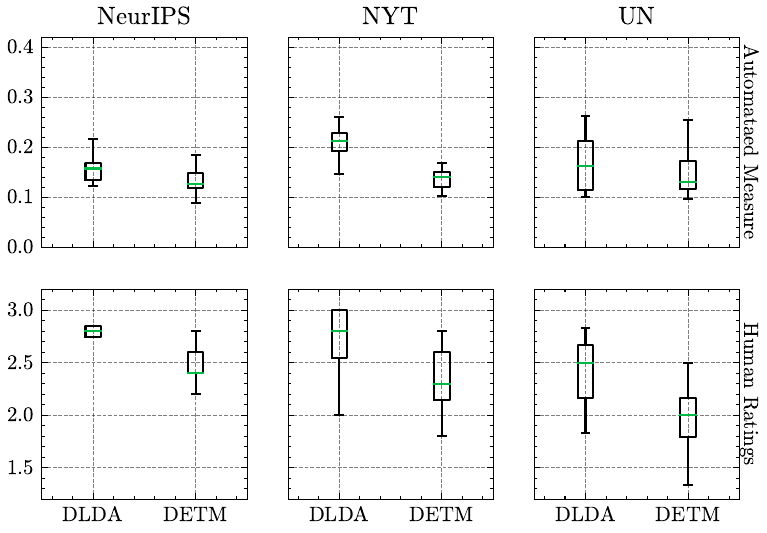}
\caption{The mean and variance for the automated measure of TTC and human evaluation results for the three datasets of (right) NeurIPS, (middle) NYT and (right) UN.}
\label{box-plot-c}
\end{figure}

\begin{figure}[hbt]
\centering
\includegraphics[scale=0.55]{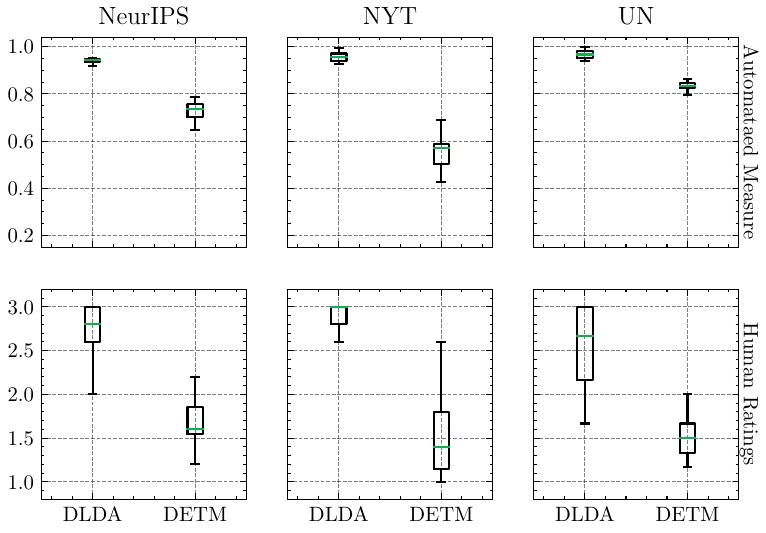}
\caption{The mean and variance for the automated measure of TTS and human evaluation results for the three datasets of (right) NeurIPS, (middle) NYT and (right) UN.}
\label{box-plot-s}
\end{figure}

In this section, we report the results of the human rating survey. We show the results of the automatic and human ratings of the randomly selected 20 topics from each model and dataset in Figure \ref{box-plot-c} and \ref{box-plot-s} for word relatedness and smoothness, respectively. The average human ratings from the survey are consistent and in line with the previous studies \cite{hoyle2021automated,roder2015exploring}.

\begin{figure}
    \centering
    \includegraphics[width=1.0\columnwidth]{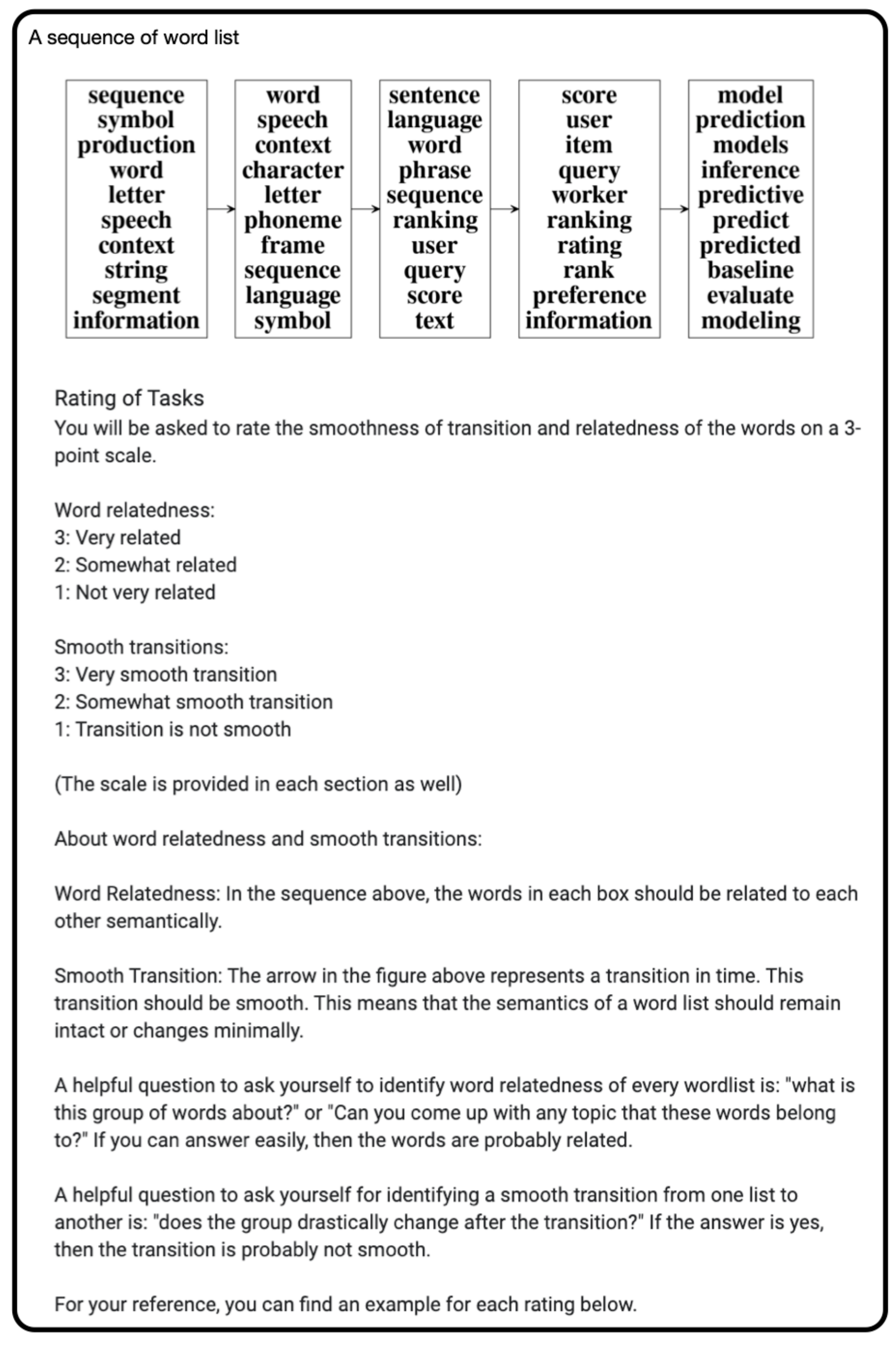}
    \caption{The instructions provided to human participants engaged in topic rating tasks. }
    \label{fig:survey_instruction}
\end{figure}

Furthermore, the instruction provided to human for rating task is shown in Figure \ref{fig:survey_instruction}. The figure depicts a sequence of words list, serving as a sample for establishing the definitions of word-relatedness and smooth transitions within the context of the study. 




\section{Word Intrusion Assessment of TTQ}
\label{sec:word-intrusion-appendix}
In this section, we continue the assessment of temporal topic quality w.r.t the intrusion levels as discussed in Section \ref{sec:intrusionAssessment}. The result is shown in Figure \ref{ttqvsnoise}. A consistent decrease in TTQ is observed for both models, with an increase in intrusion levels. This relationship is also backed by the correlations discussed in Table \ref{correlation-intrusion}. In all the cases, a strong correlation can be observed. From the intrusion task, we conclude that the TTQ measure is adequate in measuring even small changes in temporal topic quality.


\end{document}